\documentclass{article}

\usepackage{microtype}
\usepackage{graphicx}
\usepackage{subcaption}
\usepackage{booktabs} % for professional tables

\usepackage{hyperref}

\usepackage[preprint]{icml2026}

\usepackage{amsmath}
\usepackage{amssymb}
\usepackage{mathtools}
\usepackage{amsthm}

\usepackage[capitalize,noabbrev]{cleveref}

\usepackage{multirow} 
\graphicspath{{media/}}

\theoremstyle{plain}

\theoremstyle{definition}

\theoremstyle{remark}

\usepackage[textsize=tiny]{todonotes}

\icmltitlerunning{Non-Monotonic Latency in Apple MPS Decoding: KV Cache Interactions and Execution Regimes}

\begin{document}

\twocolumn[
  \icmltitle{Non-Monotonic Latency in Apple MPS Decoding:\\KV Cache Interactions and Execution Regimes}

  % It is OKAY to include author information, even for blind submissions: the
  % style file will automatically remove it for you unless you've provided
  % the [accepted] option to the icml2026 package.

  % List of affiliations: The first argument should be a (short) identifier you
  % will use later to specify author affiliations Academic affiliations
  % should list Department, University, City, Region, Country Industry
  % affiliations should list Company, City, Region, Country

  % You can specify symbols, otherwise they are numbered in order. Ideally, you
  % should not use this facility. Affiliations will be numbered in order of
  % appearance and this is the preferred way.
  \icmlsetsymbol{equal}{*}

  \begin{icmlauthorlist}
    \icmlauthor{Willy Fitra, Hendria}{equal,yyy}
    % \icmlauthor{Firstname2 Lastname2}{equal,yyy,comp}
    % \icmlauthor{Firstname3 Lastname3}{comp}
    % \icmlauthor{Firstname4 Lastname4}{sch}
    % \icmlauthor{Firstname5 Lastname5}{yyy}
    % \icmlauthor{Firstname6 Lastname6}{sch,yyy,comp}
    % \icmlauthor{Firstname7 Lastname7}{comp}
    % %\icmlauthor{}{sch}
    % \icmlauthor{Firstname8 Lastname8}{sch}
    % \icmlauthor{Firstname8 Lastname8}{yyy,comp}
    % %\icmlauthor{}{sch}
    % %\icmlauthor{}{sch}
  \end{icmlauthorlist}

  \icmlaffiliation{yyy}{Independent Researcher, Seoul, South Korea}

  \icmlcorrespondingauthor{Willy Fitra,  Hendria}{willyfitrahendria@gmail.com}

  % You may provide any keywords that you find helpful for describing your
  % paper; these are used to populate the "keywords" metadata in the PDF but
  % will not be shown in the document
  \icmlkeywords{Apple Silicon, MPS, KV cache, non-monotonic latency, execution regimes, autoregressive inference}

  \vskip 0.3in
]

% this must go after the closing bracket ] following \twocolumn[ ...

% This command actually creates the footnote in the first column listing the
% affiliations and the copyright notice. The command takes one argument, which
% is text to display at the start of the footnote. The \icmlEqualContribution
% command is standard text for equal contribution. Remove it (just {}) if you
% do not need this facility.

% Use ONE of the following lines. DO NOT remove the command.
% If you have no special notice, KEEP empty braces:
\printAffiliationsAndNotice{}  % no special notice (required even if empty)
% Or, if applicable, use the standard equal contribution text:
% \printAffiliationsAndNotice{\icmlEqualContribution}

\begin{abstract}
Autoregressive inference is typically assumed to scale predictably with decoding length, with latency increasing smoothly as generated sequence length grows. In this work, we identify unexpected non-monotonic latency behavior in the Apple MPS backend, where latency changes abruptly across nearby decoding configurations during transformer decoding. Using multiple model families (GPT-2, BLOOM, and OPT), we observe latency spikes of up to $21\times$ within specific decoding-budget intervals, followed by recovery at neighboring configurations. Controlled experiments show that these anomalies originate primarily during the decode phase rather than prefill, are not explained by memory pressure alone, and remain absent on CPU and NVIDIA CUDA backends under identical conditions. We further show that key--value (KV) cache interacts strongly with these pathological execution regimes: KV caching remains beneficial overall, but its practical speedup collapses sharply within anomalous configurations, while cache-disabled decoding still exhibits residual non-monotonic behavior. These findings suggest that autoregressive decoding on MPS enters discrete execution regimes that are not captured by coarse-grained benchmarking, highlighting the importance of hardware-aware evaluation for long-context inference.
\end{abstract}

\section{Introduction}

Autoregressive inference is commonly assumed to scale monotonically with decoding length, and key--value (KV) caching~\cite{pope2023efficiently} is widely used to accelerate autoregressive decoding~\cite{kwon2023pagedattention}. Prior work on Apple Silicon LLM inference has focused on system-level benchmarking and optimization~\cite{rajesh2025production,alizadeh2024llmflash,feng2024profiling,benazir2025profiling}, typically reporting aggregate latency or throughput measured over coarse decoding-budget sweeps or fixed evaluation points. While useful for system comparison, such evaluations often average over or sparsely sample decoding budgets, which can obscure non-local effects in decoding behavior.

However, autoregressive decoding~\cite{vaswani2017attention} differs fundamentally from batch inference. Generation proceeds token by token, and the KV cache grows dynamically with accumulated decoding context, producing evolving memory layouts and access patterns that are not captured by coarse-grained evaluations. This raises the possibility that performance may change non-smoothly across nearby decoding configurations, rather than following the monotonic trends typically assumed in system-level analyses.

A natural systems question is: \textit{when does MPS outperform CPU?} While we reproduce a model-size-dependent crossover, deeper analysis reveals that this question alone is insufficient to characterize MPS decoding behavior. Instead, we uncover a more critical phenomenon: autoregressive decoding on MPS exhibits non-monotonic latency scaling, with abrupt latency regimes emerging under specific decoding configurations and particularly pronounced under KV cache.

\begin{figure*}[t]
  \centering
  \includegraphics[width=\linewidth]{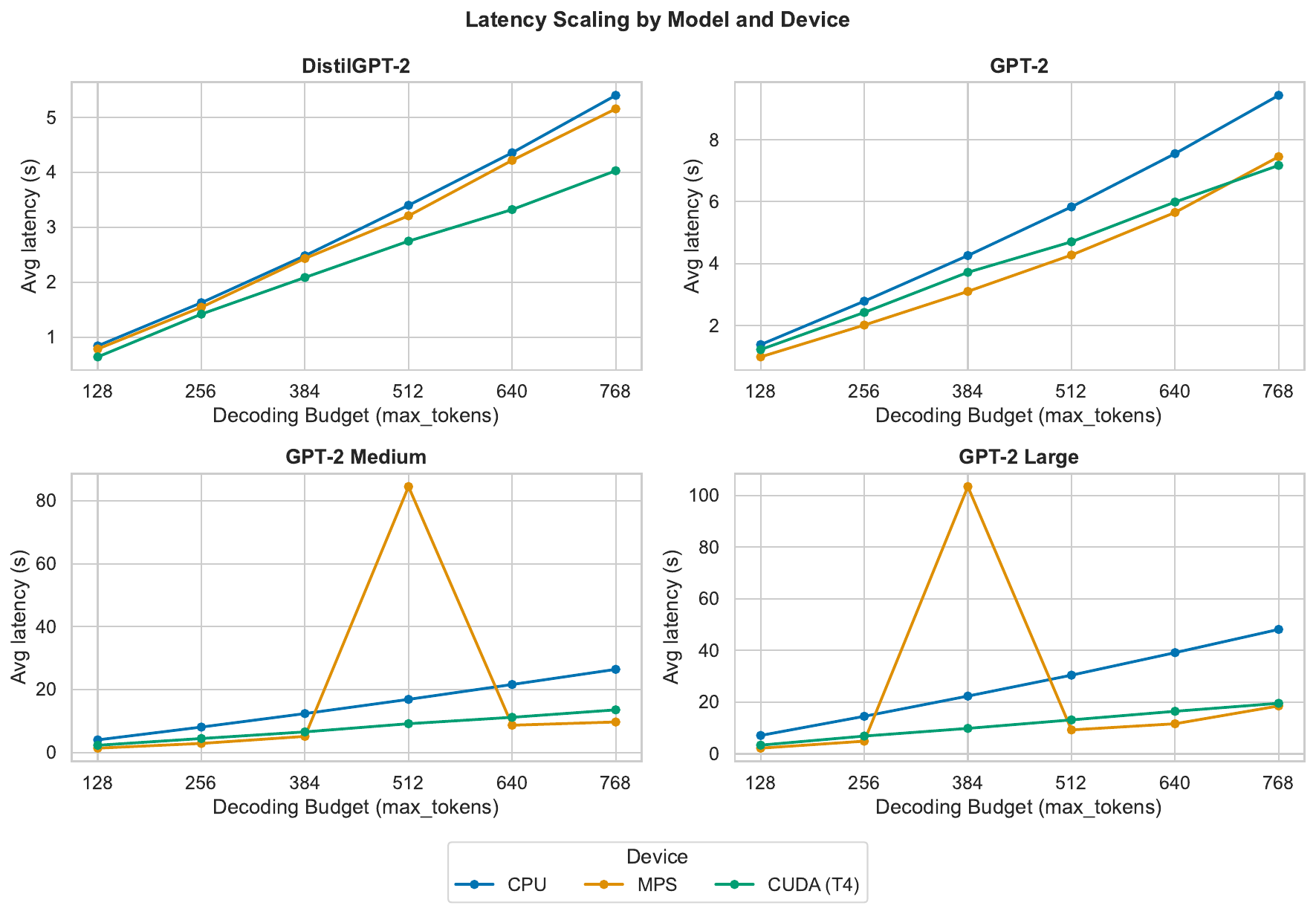}
  \caption{Latency scaling by model and backend. Each panel shows average latency versus decoding budget (\texttt{max\_tokens}) for one model. CPU and CUDA exhibit smooth monotonic scaling across all tested lengths, while MPS shows sharp non-monotonic spikes for GPT-2 Medium and GPT-2 Large at specific decoding budgets.}
  \label{fig:latency_scaling}
\end{figure*}

Figure~\ref{fig:latency_scaling} previews the observed behavior. Under MPS execution, latency can increase abruptly by more than an order of magnitude before returning to lower-latency regimes at neighboring configurations. In contrast, CPU and CUDA baselines remain smooth and monotonic under identical experimental conditions.

We find that KV cache interacts with these anomalous latency regions, affecting both the magnitude of latency spikes and the collapse of speedup at pathological decoding configurations. Through controlled experiments, we demonstrate that (1) the instability is not explained by model size or memory limits alone, (2) it emerges during decoding rather than prefill, and (3) KV cache remains nominally faster at all tested lengths, but its practical advantage is largely neutralized at pathological decoding configurations (speedup collapses from $4.9$--$20.3\times$ to $1.9\times$), while cache-off runs still show residual non-monotonicity. These findings suggest that backend execution behavior contributes to performance variations in specific regimes that are not captured by coarse benchmarks.

\paragraph{Contributions.}
\begin{itemize}
\item We identify non-monotonic latency behavior in MPS autoregressive decoding, with spikes up to $21\times$ at model-dependent decoding budgets, while CPU and CUDA T4 baselines remain strictly monotonic under identical conditions.

\item We demonstrate that this behavior generalizes across multiple model families (GPT-2, BLOOM, and OPT), indicating a backend-level phenomenon rather than model-specific effects.

\item We demonstrate sharp regime transitions in MPS decoding, where latency jumps abruptly within narrow decoding-budget intervals despite substantially smoother observed memory scaling.

\item We show that KV cache highlights execution sensitivity, revealing a speedup reduction (from $4.9$--$20.3\times$ to $1.9\times$) at pathological decoding configurations, while not removing the underlying non-monotonic structure.

\item We analyze the sensitivity of the anomalous regime to prompt length and decoding budget, showing that nearby decoding configurations can exhibit sharply different latency behavior under MPS decoding.
\end{itemize}

\section{Related Work}

\subsection{Attention Efficiency and Sequence-Length Optimization}

A large body of work focuses on improving the efficiency of transformer attention mechanisms under long-context inference. KV cache reduces recomputation during autoregressive decoding while increasing memory usage with accumulated decoding context~\cite{pope2023efficiently,kwon2023pagedattention,zhang2023h2o,willette2025training}. Complementary kernel-level optimizations such as FlashAttention~\cite{dao2022flashattention} improve efficiency through memory-aware attention computation and IO-aware kernel design.

While these methods significantly improve efficiency, they generally assume monotonic cost growth with decoding length. That is, latency is expected to increase smoothly as decoding budget increases, reflecting predictable memory and compute scaling in modern accelerators.

In contrast, our results show that this assumption does not hold on the Apple MPS backend, where autoregressive decoding exhibits non-monotonic latency behavior.

\subsection{Language Model on Apple Silicon}

Recent work has explored production-grade runtime comparisons across local inference stacks~\cite{rajesh2025production}, alongside weight streaming approaches~\cite{alizadeh2024llmflash}, KV-cache optimization for edge inference~\cite{horton2024kvprediction}, and profiling studies of Apple Silicon performance~\cite{feng2024profiling,benazir2025profiling}. While these studies provide valuable system-level insights, they primarily report aggregate latency or throughput under coarse decoding-budget sweeps.

The PyTorch MPS backend~\cite{pytorch2022mps} enables GPU inference via Metal Performance Shaders~\cite{apple2026mps}, but its internal execution behavior is not directly exposed through the PyTorch API, making performance characteristics difficult to fully model or predict.

To the best of our knowledge, existing evaluations do not systematically examine fine-grained decoding-budget variations or potential non-monotonic latency regimes in autoregressive decoding. This motivates our study of MPS execution behavior under controlled decoding settings.

\section{Experimental Setup}

Unless otherwise stated, all experiments follow the default configuration described in this section. Any deviations, such as changes in hardware, instrumentation, or ablation settings, are explicitly specified in the relevant subsection and are not mixed across experiments.

\subsection{Experimental Configuration}

Experiments are conducted on a MacBook Pro equipped with an Apple M3 Max processor (14-core CPU, 30-core GPU, 36~GB unified memory) running macOS 14.8.1 (Sonoma). The primary software stack includes Python~3.9.6, PyTorch~2.8.0~\cite{paszke2019pytorch}, HuggingFace Transformers~4.57.6~\cite{wolf2020transformers}, and LitServe~0.2.16~\cite{lightningai2026litserve}, using the MPS backend for GPU-accelerated execution. To assess backend stability, we validated that qualitatively similar non-monotonic behavior persists across multiple PyTorch versions (2.7.0, 2.8.0, and 2.11.0), indicating the phenomenon is not tied to a specific release. Key findings were independently reproduced on an Apple M3 Pro (11-core CPU, 8-core GPU, 18~GB unified memory) under identical conditions, demonstrating the cross-device generality of the observed phenomena.

We evaluate transformer language models across multiple families, including GPT-2~\cite{radford2019gpt2}, BLOOM~\cite{scao2022bloom}, and OPT~\cite{zhang2022opt}, spanning a range of model sizes and architectures. Inference is also tested across three execution backends: CPU, Apple MPS (Metal Performance Shaders), and NVIDIA CUDA (T4)~\cite{nvidia_t4_2018}, under identical configurations.

\subsection{Benchmark Design}

All experiments are served through LitServe~\cite{lightningai2026litserve}, a production-grade Python inference serving framework.
Using a real serving stack rather than bare model calls ensures that overheads from request handling, worker scheduling, and response serialization are present throughout, making the observed anomalies relevant to actual deployment scenarios.
Crucially, the anomalies manifest at the decode phase (token-level timing shows prefill is negligible), which is decoupled from LitServe's request handling, indicating the instability originates in the MPS backend's kernel execution rather than framework scheduling.

Each experiment runs in an isolated server process with 3 warmup runs (discarded) and 5 measured runs (averaged). We use deterministic decoding with a fixed seed and apply token-level timing for selected experiments. Decoding budgets (\texttt{max\_tokens}) range from 128 to 768 tokens. We disable early stopping based on EOS tokens and enforce a fixed decoding budget (\texttt{max\_tokens}), allowing EOS tokens to be generated but not treated as stopping criteria. This isolates backend execution dynamics from model-dependent stopping behavior and ensures strict comparability across runs by preventing stochastic truncation from affecting sequence-length-dependent latency measurements. Unless otherwise specified, references to decoding budget in plots and tables correspond to the configured \texttt{max\_tokens} value rather than realized output length. Because EOS-based stopping is disabled, the realized generated output length always equals the configured decoding budget in our experiments. We use "generated tokens" to refer to the realized generated output length and "total sequence length" to refer to prompt tokens plus generated tokens.

\begin{table*}[t]
\centering
\caption{Latency collapse on MPS at specific decoding budgets. The anomalous configuration performs substantially worse than both shorter and longer neighboring decoding configurations, demonstrating strictly non-monotonic scaling behavior.}
\label{tab:anomaly_summary}

\begin{tabular}{l | c | c | c}
\toprule
Model & Decoding Budget & Latency & Throughput (tok/s) \\
\midrule

\multirow{3}{*}{GPT-2 Medium} 
& 384 & 5.12 $\pm$ 0.06 & 75.27 $\pm$ 0.83 \\
& \textbf{512} & \textbf{84.54 $\pm$ 12.18} & \textbf{6.20 $\pm$ 0.99} \\
& 640 & 8.67 $\pm$ 0.05 & 73.88 $\pm$ 0.40 \\

\midrule

\multirow{3}{*}{GPT-2 Large} 
& 256 & 4.91 $\pm$ 0.02 & 52.12 $\pm$ 0.24 \\
& \textbf{384} & \textbf{103.41 $\pm$ 4.99} & \textbf{3.70 $\pm$ 0.18} \\
& 512 & 9.27 $\pm$ 0.05 & 55.20 $\pm$ 0.30 \\

\bottomrule
\end{tabular}
\end{table*}

\subsubsection{Measurement methodology.}

The main experiments (Section~\ref{sec:main}) measure end-to-end latency without instrumentation. For per-token profiling experiments (KV cache ablation and instability probe), a synchronization barrier is inserted after each decode step to obtain accurate timing. This changes the execution behavior relative to the baseline setting and may slightly affect the measured latency. Consequently, absolute latency values from instrumented and uninstrumented experiments are not directly comparable. All KV-cache-enabled and disabled comparisons use identical instrumentation and remain internally consistent. Importantly, the non-monotonic behavior is already present in uninstrumented end-to-end measurements; per-step instrumentation is used only to attribute latency to prefill versus decode phases.

\section{Results}

\subsection{Emergence of Non-Monotonic Scaling}
\label{sec:main}

\begin{table}[t]
\centering
\caption{Average latency (s) across all tested decoding budgets for each model and backend.}
\label{tab:summary_cross_backend}
\begin{tabular}{l | r | r | r}
\toprule
Model & CPU & CUDA & MPS \\
\midrule
DistilGPT-2 & 3.018 & 2.374 & 2.890 \\
GPT-2 & 5.208 & 4.205 & 3.914 \\
GPT-2 Medium & 14.899 & 7.862 & 18.713 \\
GPT-2 Large & 26.966 & 11.555 & 25.001 \\
\bottomrule
\end{tabular}
\end{table}

Table~\ref{tab:summary_cross_backend} summarizes average latency across all tested decoding budgets for CPU, CUDA, and MPS backends. Across all backends, mean latency increases with model size, reflecting the expected increase in computational cost for larger transformer models. CUDA consistently outperforms CPU, while MPS performance is more variable: although competitive for smaller models, MPS becomes comparable to or slower than CPU for GPT-2 Medium and GPT-2 Large.

Although the aggregate averages suggest reduced MPS efficiency for larger models, they do not reveal the underlying cause of the slowdown. Figure~\ref{fig:latency_scaling} shows that the slowdown is not due to uniformly poor scaling. Instead, MPS exhibits sharp latency spikes at specific decoding budgets for GPT-2 Medium and GPT-2 Large, while CPU and CUDA remain smooth and monotonic throughout. Averaging latency across decoding budgets therefore obscures the presence of discrete pathological execution regimes.

As summarized in Table~\ref{tab:anomaly_summary}, the anomalous behavior is strongly non-monotonic and model-size-dependent. The anomalous configurations perform substantially worse than both shorter and longer neighboring decoding configurations, demonstrating strictly non-monotonic scaling behavior. For GPT-2 large, the slowdown even reaches up to $21\times$ relative to adjacent non-anomalous configurations. Additionally, entry into high-latency regime is model-size-dependent, with larger models entering the degraded regime at smaller decoding budgets. This suggests that the triggering condition is also affected by model size, rather than by the decoding budget alone.

\subsection{Cross-Device Anomaly}

\begin{figure*}[t]
  \centering
  \includegraphics[width=\linewidth]{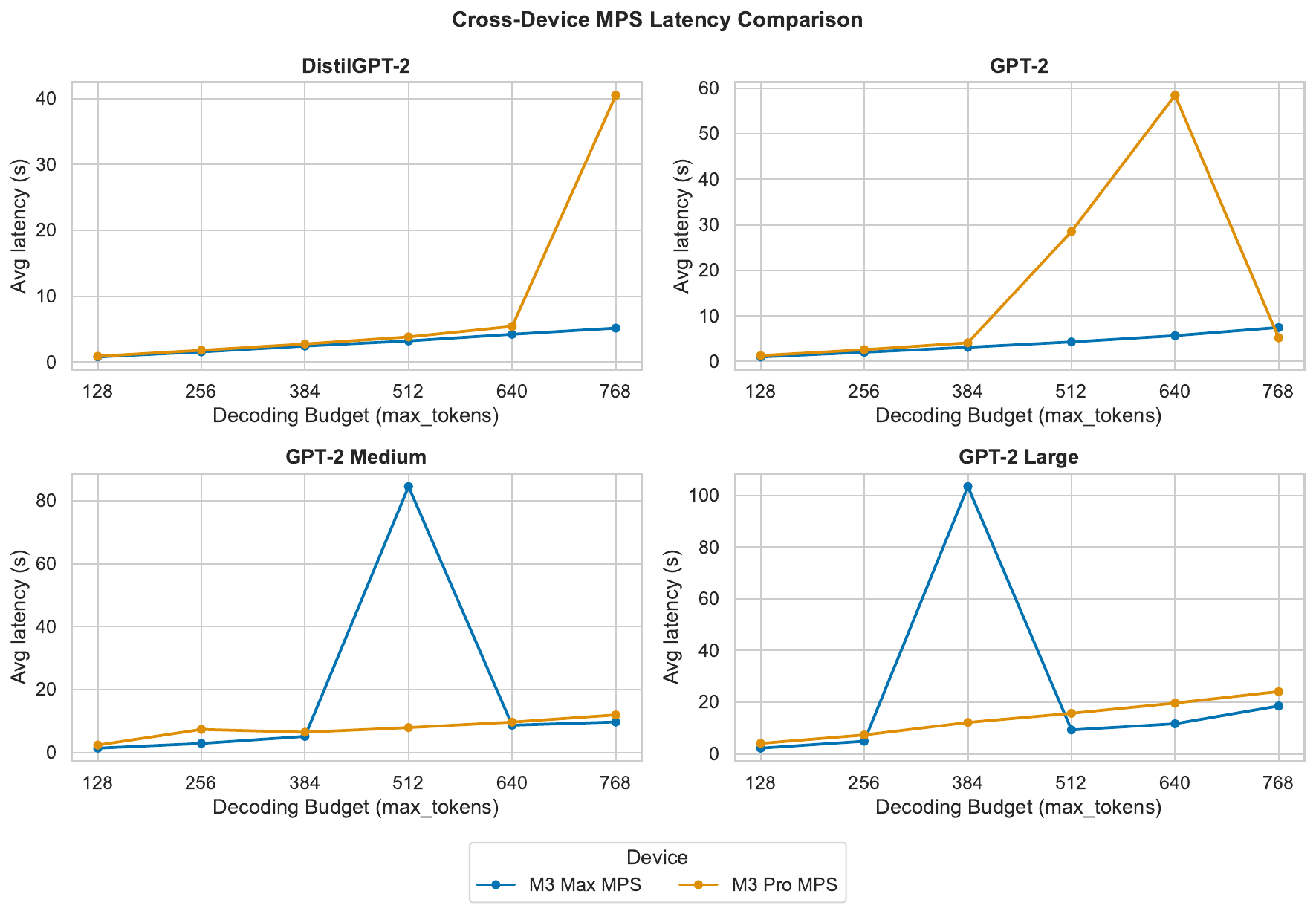}
  \caption{Cross-device MPS latency comparison between M3 Max and M3 Pro. Each panel shows latency versus decoding budget (\texttt{max\_tokens}) for one model under MPS execution. The anomaly boundary shifts systematically across devices and model sizes: M3 Max exhibits spikes for larger models at smaller decoding budgets, while M3 Pro exhibits instability even for smaller models.}
  \label{fig:cross_device_mps}
\end{figure*}

The phenomenon is also observed across multiple Apple Silicon configurations. Figure~\ref{fig:cross_device_mps} compares latency scaling on M3 Max and M3 Pro devices under identical experimental conditions. The anomaly boundary differs across hardware configurations: the higher-capacity M3 Max exhibits instability primarily for GPT-2 Medium and GPT-2 Large, whereas the lower-capacity M3 Pro shows anomalies even for smaller models such as GPT-2 and DistilGPT-2~\cite{sanh2019distilbert}. 

This shift toward smaller models on the lower-capacity device suggests that the triggering boundary depends on hardware capacity and backend execution characteristics rather than model architecture alone.

\subsection{Instability Probe}

\begin{figure}[t]
  \centering
  \includegraphics[width=\linewidth]{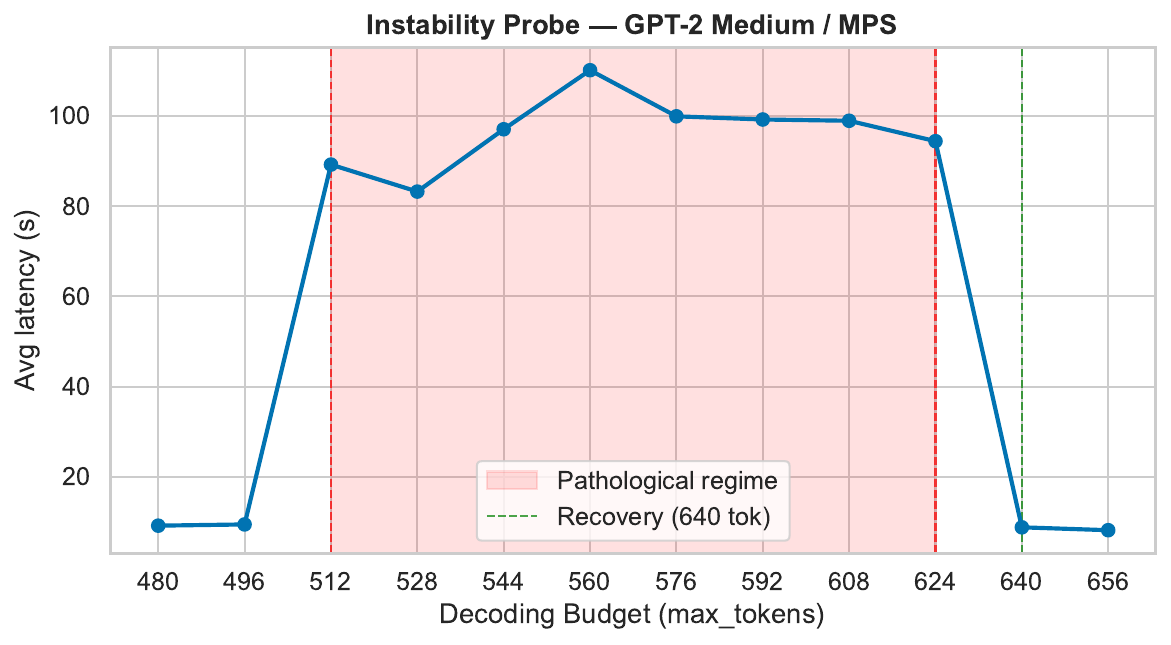}
  \caption{Instability probe: average latency vs.\ \texttt{max\_tokens} for GPT-2 Medium on MPS at 16-token intervals. The shaded region marks the anomalous regime (512--624 tokens). Dashed lines indicate the onset and recovery boundaries.}
  \label{fig:instability_probe}
\end{figure}

The transition into the anomalous regime is abrupt rather than gradual, as illustrated in Figure~\ref{fig:instability_probe}. A focused instability sweep over 480--656 tokens at 16-token spacing reveals abrupt transitions between low- and high-latency regions. Latency remains within the normal range at 480--496 tokens (9.2--9.5s), then jumps abruptly at 512 tokens (89.2s). The degraded regime persists through 624 tokens (83.2--110.1s across sampled points), before collapsing back to normal at 640--656 tokens (8.8s and 8.2s, respectively).

These measurements show that the instability is not an isolated outlier at 512 tokens, but a distinct execution regime spanning multiple neighboring decoding configurations. Across the 5 measured runs, latency within the degraded regime remains consistently high, with standard deviation ranging from 8--14~s at 512 tokens. This indicates stable but pathological execution behavior rather than transient measurement noise. The regime transition occurs within a single 16-token step on both sides (entry into higher-latency between 496 and 512, return to lower-latency between 624 and 640), which is inconsistent with smooth scaling and instead suggests a discrete change in backend execution behavior.

Per-step timing in the instrumented runs confirms that prefill cost is negligible throughout. At 512 tokens, the prefill phase completes in 0.11~s (0.1\% of total latency), while at the adjacent normal point (496 tokens) prefill is 0.08~s (0.8\% of total latency). The latency difference between 496 tokens (9.5~s) and 512 tokens (89.2~s) is therefore attributable almost entirely to the decode phase (9.4~s vs.\ 89.1~s respectively), rather than changes in prefill cost.

\subsection{KV Cache Ablation}
\label{sec:kvcache}

We perform a controlled ablation on GPT-2 Medium using MPS, varying KV cache state across decoding budgets from 128--768 tokens. All other conditions (model, prompt, seed, and instrumentation) are held constant. The KV-cache-disabled condition uses the same protocol as the main experiments (3 warmup runs and 5 measured runs).

\begin{figure}[t]
\centering
\includegraphics[width=\linewidth]{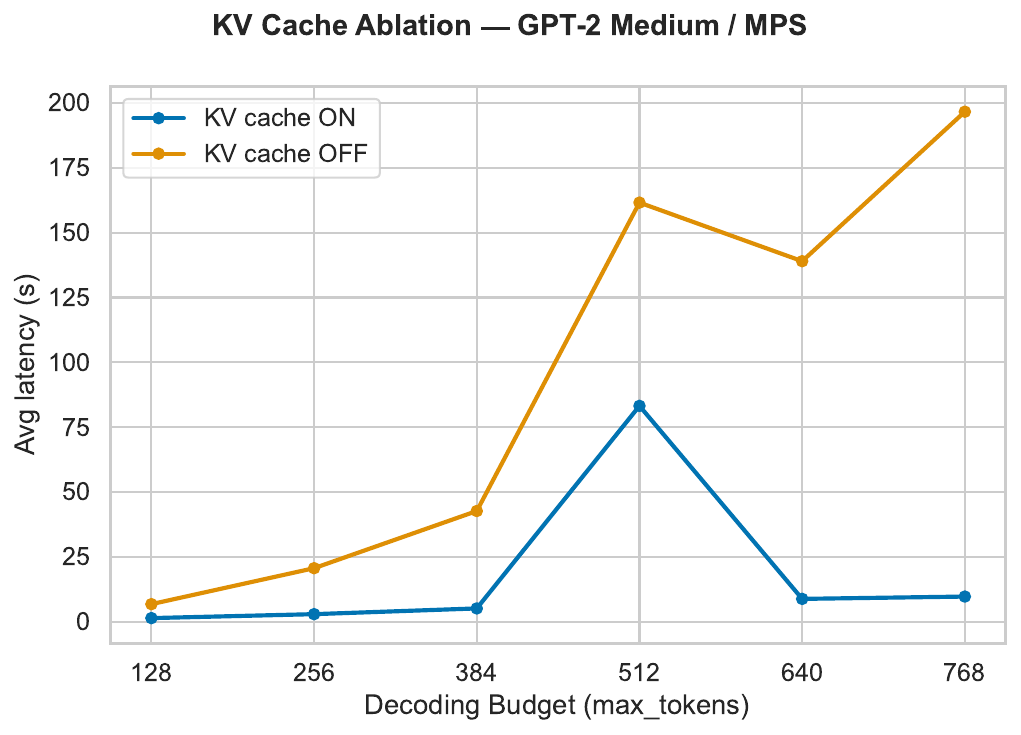}
\caption{KV cache ablation for GPT-2 Medium on MPS. With KV cache enabled, latency exhibits a sharp non-monotonic spike at decoding budget 512. With KV cache disabled, latency is higher throughout and remains non-monotonic. Together, these results indicate that KV caching accentuates the observed latency discontinuity while still providing substantial speedup outside pathological decoding configurations.}
\label{fig:kv_cache_ablation}
\end{figure}

With KV cache enabled, latency exhibits a sharp anomaly at 512 tokens, increasing from 5.1s at 384 tokens to 83.2s before recovering to 8.8s at 640 tokens, as shown in Figure~\ref{fig:kv_cache_ablation}. When KV cache is disabled, non-monotonic behavior remains present: latency reaches 161.6s at 512 tokens and 139.0s at 640 tokens. Thus, disabling KV cache does not eliminate the anomalous regime.

However, KV cache substantially changes the severity of the anomaly. At non-anomalous decoding configurations, cache-disabled execution is $4.9\times$--$20.3\times$ slower than cache-enabled execution, as summarized in Table~\ref{tab:kv_ablation}. At the anomalous 512-token configuration, this ratio collapses to only $1.9\times$ (161.6s vs.\ 83.2s), despite cache-disabled decoding requiring substantially more computation. This indicates that the practical speed advantage of KV caching is substantially reduced within the anomalous regime.

Across runs, variability also increases at longer cache-disabled decoding budgets (e.g., standard deviation 21.5s at 640 tokens and 40.2s at 768 tokens), likely reflecting the substantially larger computational cost of cache-disabled decoding at long contexts. Generated outputs were observed to remain identical across KV-cache-enabled and disabled configurations under deterministic greedy decoding with fixed seeds for the tested decoding budgets, indicating that the observed differences are associated with execution behavior rather than output divergence in these experiments.

\begin{table}[t]
\centering
\caption{KV cache ablation for GPT-2 Medium on MPS: latency (s) with cache enabled vs.\ disabled. The \textbf{bold row} marks the anomalous configuration (512 tokens). The ratio (cache-off / cache-on) shows that the practical speedup from KV caching collapses at the anomalous configuration.}
\label{tab:kv_ablation}
\begin{tabular}{r | r | r | r}
\toprule
\shortstack{Decoding\\Budget} & Cache ON (s) & Cache OFF (s) & Ratio \\
\midrule
128 & 1.38 $\pm$ 0.02 & 6.76 $\pm$ 0.10 & $4.9\times$ \\
256 & 2.90 $\pm$ 0.03 & 20.63 $\pm$ 0.79 & $7.1\times$ \\
384 & 5.11 $\pm$ 0.03 & 42.67 $\pm$ 7.69 & $8.4\times$ \\
\textbf{512} & \textbf{83.21 $\pm$ 7.98} & \textbf{161.60 $\pm$ 1.14} & $1.9\times$ \\
640 & 8.79 $\pm$ 0.04 & 139.00 $\pm$ 21.48 & $15.8\times$ \\
768 & 9.69 $\pm$ 0.04 & 196.64 $\pm$ 40.22 & $20.3\times$ \\
\bottomrule
\end{tabular}
\end{table}

\subsection{Memory Pressure Analysis}
To investigate whether the instability can be explained by memory pressure, we profile peak memory usage across both GPU and system levels. For GPU-allocated memory on MPS, peak allocation increases monotonically and approximately linearly across all decoding budgets, from 1426\ MB at 128 tokens to 1550\ MB at 768 tokens, without any sharp transition at the anomalous 512-token boundary.

At the system level, peak unified-memory usage increases substantially near the anomalous regime, rising from 13~GB at 128 tokens to 33~GB at 512 tokens. However, system memory usage does not track the latency anomaly itself: elevated memory usage persists beyond the anomalous region, including at 640 tokens where latency has already returned to the normal regime. This suggests that memory growth alone is insufficient to explain the abrupt onset and recovery of the latency anomaly.

Table~\ref{tab:memory_unified} reports both measurements for GPT-2 Medium, the model exhibiting the clearest 512-token instability. The smooth GPU-memory scaling and the mismatch between system-memory behavior and recovery are more consistent with execution-regime effects than simple memory-capacity exhaustion.

\begin{table}[t]
\centering
\caption{Memory usage for GPT-2 Medium across GPU and system levels on M3 Max. MPS-allocated memory increases approximately linearly across decoding budgets, while system memory rises substantially near the anomalous region but remains elevated after latency returns to the lower-latency regime.}
\label{tab:memory_unified}
\begin{tabular}{r | r | r}
\toprule
\shortstack{Decoding \\ Budget} & \shortstack{Peak MPS \\ (MB)} & \shortstack{Peak System \\ (MB)} \\
\midrule
128 & 1426 & 13,166 \\
256 & 1451 & 16,809 \\
384 & 1476 & 22,111 \\
\textbf{512} & \textbf{1500} & \textbf{32,881} \\
640 & 1525 & 33,261 \\
768 & 1550 & 31,227 \\
\bottomrule
\end{tabular}
\end{table}

\subsection{Cross-Model Generalization}

To evaluate whether the observed non-monotonic latency behavior is specific to GPT-2 architectures, we evaluate two additional model families: BLOOM-560M~\cite{scao2022bloom} and OPT-350M~\cite{zhang2022opt}, under identical MPS execution settings.

Table~\ref{tab:cross_model} summarizes latency across decoding budgets. Both models exhibit the same qualitative non-monotonic structure: stable low-latency scaling at short decoding budgets, sharp degradation at intermediate budgets, and recovery at longer sequences. While the exact transition points differ across architectures, the regime structure is consistent.

Quantitatively, OPT-350M shows more than an order-of-magnitude latency increase between 384 and 512 tokens, and BLOOM-560M exhibits elevated variance in the anomalous region (up to $\sim 5$--$6$ seconds standard deviation), while non-anomalous regions remain stable.

These results indicate that the non-monotonic execution regimes are not specific to GPT-style models, but generalize across diverse transformer architectures, supporting a backend-level origin in MPS execution.

\begin{table}[t]
\centering
\caption{Cross-model evaluation on BLOOM-560M and OPT-350M under the MPS backend. Latency is reported as mean $\pm$ std.}
\label{tab:cross_model}

\begin{tabular}{l | c | c}
\toprule
Model & Decoding Budget & Avg Latency (s) \\
\midrule

\multirow{6}{*}{\texttt{BLOOM-560M}} 
& 128 & 2.90 $\pm$ 0.32 \\
& 256 & 7.53 $\pm$ 0.31 \\
& 384 & 15.17 $\pm$ 0.05 \\
& 512 & 80.76 $\pm$ 5.04 \\
& 640 & 134.58 $\pm$ 6.13 \\
& 768 & 40.80 $\pm$ 0.21 \\

\midrule

\multirow{6}{*}{\texttt{OPT-350M}} 
& 128 & 1.71 $\pm$ 0.01 \\
& 256 & 3.74 $\pm$ 0.04 \\
& 384 & 7.40 $\pm$ 0.07 \\
& 512 & 95.13 $\pm$ 2.03 \\
& 640 & 15.71 $\pm$ 0.13 \\
& 768 & 19.35 $\pm$ 0.11 \\

\bottomrule
\end{tabular}
\end{table}

\subsection{Prompt-Length Analysis}
\label{sec:prompt}

We use representative short and long prompts rather than exhaustively sweeping prompt lengths, focusing on boundary sensitivity rather than full prompt-length scaling.

To test whether the instability is governed primarily by the decoding budget (\texttt{max\_tokens}) or by total sequence length (prompt tokens + generated tokens), we hold \texttt{max\_tokens} constant and vary the input prompt length across two conditions: a short prompt of 3 tokens and a long prompt of approximately 65 tokens, resulting in total sequence lengths that differ by approximately 62 tokens at each configuration.

\begin{figure}[t]
  \centering
  \includegraphics[width=\linewidth]{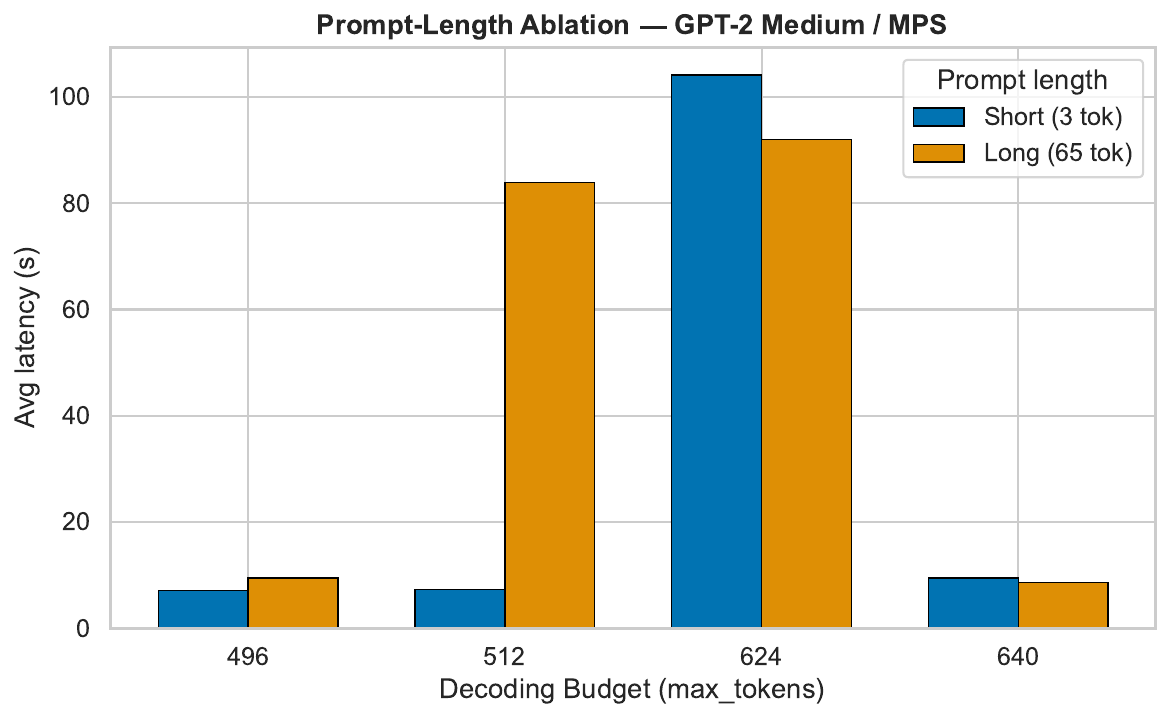}
  \caption{Prompt-length ablation for GPT-2 Medium on MPS. At \texttt{max\_tokens}~=~512 the long prompt (total $\approx$577 tokens) enters the anomalous execution regime while the short prompt (total $\approx$515 tokens) remains normal, consistent with an influence of total sequence length on onset behavior.}
  \label{fig:prompt_ablation}
\end{figure}

Figure~\ref{fig:prompt_ablation} illustrates the resulting latency behavior. The results suggest that entry into the high-latency region is associated with both budget and total sequence length. As also shown in Table~\ref{tab:prompt_ablation}, the long prompt enters the high-latency region at \texttt{max\_tokens} = 512 (83.9s) while the short prompt remains in the low-latency regime (7.3s), consistent with the long prompt crossing an effective total-length threshold earlier. However, the relationship is not strictly monotonic within the high-latency region, indicating that total sequence length alone does not fully explain runtime behavior.

Return to lower latency occurs at the same decoding budget (\texttt{max\_tokens}) configuration for both prompt conditions, despite differing total sequence lengths. This separation between entry sensitivity (associated with accumulated sequence size during decoding) and return to lower-latency behavior (associated with the configured decoding budget) suggests an interaction between prompt-dependent and configuration-dependent factors, though the exact mechanism cannot be determined from user-level observations.

The asymmetry between entry and recovery behavior suggests that transitions into and out of the high-latency region may not follow the same dependency structure.

\begin{table*}[t]
\centering
\caption{Prompt-length ablation results for GPT-2 Medium on MPS. Short prompt: 3 tokens; long prompt: 65 tokens. Total sequence length = \texttt{max\_tokens} + prompt tokens.}
\label{tab:prompt_ablation}
\begin{tabular}{r | r | r | r | r}
\toprule
& \multicolumn{2}{c}{Short prompt (3 tok)} & \multicolumn{2}{c}{Long prompt (65 tok)} \\
\cmidrule(lr){2-3}\cmidrule(lr){4-5}
Decoding Budget & Total seq & Latency (s) & Total seq & Latency (s) \\
\midrule
496 & $\approx$499 & 7.1 & $\approx$561 & 9.5 \\
512 & $\approx$515 & 7.3 & $\approx$577 & 83.9 \\
624 & $\approx$627 & 104.2 & $\approx$689 & 92.0 \\
640 & $\approx$643 & 9.5 & $\approx$705 & 8.6 \\
\bottomrule
\end{tabular}
\end{table*}

\section{Discussion}

Across all experiments, a consistent pattern emerges: MPS autoregressive decoding does not scale smoothly with increasing decoding budgets (\texttt{max\_tokens}), but instead exhibits abrupt non-monotonic latency transitions across nearby decoding configurations. These regimes are jointly influenced by model size, KV cache state, and total sequence length, but are not fully explained by any single factor in isolation. Instead, the observations are consistent with regime shifts in execution behavior, where decoding state size appears to influence internal runtime decisions such as scheduling or kernel selection within the MPS backend.

\subsection{Implications for Practice}

These findings have direct consequences for practitioners deploying autoregressive models on Apple Silicon.

First, KV cache does not provide consistent speedup on MPS across all decoding budgets under the tested settings. In particular, its benefit significantly degrades within the anomalous latency regime, where the relative speedup over cache-disabled execution collapses compared to neighboring configurations.

Second, benchmark methodology is critical. Coarse measurements at a single sequence length may fail to reveal these effects entirely, as the instability only emerges beyond model- and configuration-dependent decoding-budget thresholds. As a result, reporting performance at a single decoding budget, or aggregating evaluations across a narrow set of decoding budgets, may provide an incomplete view of the underlying non-monotonic performance behavior.

\subsection{Limitations}

Core characterization and initial discovery were conducted on M3 Max; qualitatively similar non-monotonic behavior was independently reproduced on M3 Pro under identical conditions, suggesting that the phenomenon is not isolated to a single Apple Silicon configuration. We also observe similar behavior across multiple PyTorch versions (2.7.0, 2.8.0, and 2.11.0), indicating that the anomaly is unlikely to be tied to a specific software release. Reproducing the experiments on additional Apple Silicon generations (e.g., M1, M2) and across broader decoding configurations would further strengthen generalizability, which we leave to future work.

Additionally, the MPS backend does not expose kernel-level instrumentation through the PyTorch API, limiting visibility into low-level execution behavior from user space. Prior work has shown that understanding GPU execution behavior often requires low-level profiling or microbenchmarking~\cite{jia2019t4microbench}. Such instrumentation is not exposed through the PyTorch MPS execution stack. We leave deeper investigation of the underlying mechanism and potential mitigation strategies to future work.

\section{Conclusion}

We presented evidence that autoregressive inference on Apple MPS exhibits non-monotonic latency behavior across decoding budgets, characterized by abrupt regime shifts rather than smooth scaling. This behavior is consistent across multiple model families (GPT-2, BLOOM, OPT) and persists under controlled experimental conditions. Our results further show that KV cache interacts with these regimes, affecting the severity and relative speedup of the observed anomalies, and that neither memory usage nor model architecture alone fully explains the observed behavior. Overall, these findings show that MPS execution introduces strong, non-monotonic performance regimes with latency spikes of up to an order of magnitude, which are not captured by coarse-grained benchmarking practices.

\section*{Code Availability}

The benchmark harness and configuration files used in this study will be released upon publication.

\section*{Impact Statement}

This paper presents work whose goal is to advance the field of Machine Learning. There are many potential societal consequences of our work, none which we feel must be specifically highlighted here.

% In the unusual situation where you want a paper to appear in the
% references without citing it in the main text, use \nocite
\nocite{langley00}

\bibliography{references}
\bibliographystyle{icml2026}

\end{document}